\def\year{2022}\relax
\documentclass[letterpaper]{article} 
\usepackage{aaai22}  
\usepackage{times}  
\usepackage{helvet}  
\usepackage{courier}  
\usepackage[hyphens]{url}  
\usepackage{graphicx} 
\usepackage{natbib}  
\usepackage{caption} 
\DeclareCaptionStyle{ruled}{labelfont=normalfont,labelsep=colon,strut=off} 
\usepackage{algorithm}
\usepackage{algorithmic}

\usepackage{amsmath}
\usepackage{amsfonts}
\usepackage{amssymb}
\usepackage{overpic}
\usepackage{multirow}
\usepackage[table,xcdraw]{xcolor}
\usepackage{mathtools}
\usepackage{makecell}
\usepackage{lscape}
\usepackage{color}
\usepackage{xcolor}
\usepackage{hyperref}

\hypersetup{
    colorlinks=false,  
    citebordercolor={0 1 0},
    pdfborderstyle={/S/S/W 1},
}

\usepackage{newfloat}
\usepackage{listings}
\floatstyle{ruled}
\newfloat{listing}{tb}{lst}{}
\floatname{listing}{Listing}

\title{Gradual Vigilance and Interval Communication: Enhancing Value Alignment in Multi-Agent Debates}
\author {
    Rui Zou\textsuperscript{\rm 1\rm 2},
    Mengqi Wei\textsuperscript{\rm 1\rm 2},
    Jintian Feng\textsuperscript{\rm 1\rm 2},
    Qian Wan \textsuperscript{\rm 1 \rm 2},
    Jianwen Sun\textsuperscript{\rm 1\rm 2},
    Sannyuya Liu\thanks{Corresponding author.} \textsuperscript{\rm 1\rm 2}
}
\affiliations {
    \textsuperscript{\rm 1} National Engineering Research Center of Educational Big Data, Central China Normal University, Wuhan 430079, China\\
    \textsuperscript{\rm 2} Faculty of Artificial Intelligence in Education, Central China Normal University, Wuhan 430079, China\\
    zouruixyz@mails.ccnu.edu.cn
}

\usepackage{bibentry}

\graphicspath{{figures/}}

\begin{document}

\renewcommand{\dblfloatpagefraction}{.9}

\maketitle



\begin{abstract}
In recent years, large language models have shown exceptional performance in fulfilling diverse human needs. However, their training data can introduce harmful content, underscoring the necessity for robust value alignment. Mainstream methods, which depend on feedback learning and supervised training, are resource-intensive and may constrain the full potential of the models. Multi-Agent Debate (MAD) offers a more efficient and innovative solution by enabling the generation of reliable answers through agent interactions.
To apply MAD to value alignment, we examine the relationship between the helpfulness and harmlessness of debate outcomes and individual responses, and propose a MAD-based framework—Gradual Vigilance and Interval Communication (GVIC). GVIC allows agents to assess risks with varying levels of vigilance and to exchange diverse information through interval communication. We theoretically prove that GVIC optimizes debate efficiency while reducing communication overhead.
Experimental results demonstrate that GVIC consistently outperforms baseline methods across various tasks and datasets, particularly excelling in harmfulness mitigation and fraud prevention. Additionally, GVIC exhibits strong adaptability across different base model sizes, including both unaligned and aligned models, and across various task types.
\end{abstract}

\section{Introduction}\label{sec:introduction}

In recent years, large language models (LLM) have shown exceptional capabilities in addressing diverse human information needs \cite{brown2020language,bubeck2023sparks,touvron2023llama,wu2023brief}. However, the diverse nature of training data inevitably exposes LLM to misleading, harmful, and toxic content \cite{bai2022constitutional,ouyang2022training}, highlighting the critical importance of aligning these models with human values to mitigate potential negative impacts \cite{anwar2024foundational,barrett2023identifying}.

Current alignment methods predominantly rely on external supervision or supervised fine-tuning. For example, Reinforcement Learning from Human Feedback (RLHF) \cite{achiam2023gpt,dai2023safe} requires extensive human preference data and a two-stage training process, resulting in considerable costs. Other approaches, such as RLAiF \cite{lee2023rlaif}, aim to reduce human supervision but still rely on advanced LLM, which are resource-intensive. Supervised fine-tuning (SFT) methods, including Aligner \cite{ji2024aligner} and PRO \cite{song2024preference}, optimize LLM using preference loss functions, but these methods often constrain the model's ability to surpass human performance.
Unlike these approaches, \cite{du2023improving} demonstrates that Multi-Agent Debate (MAD) enhances resource efficiency and creativity through interactions among multiple LLM. In MAD, agents collaboratively propose and refine responses, sharing both answers and historical data to jointly complete tasks. \cite{singhal2023towards} introduced a method that integrates multiple reasoning paths, optimizing answers through multi-round aggregation. Other approaches include one-on-one debates, synchronous discussions, summarizer-aided discussions \cite{chan2023chateval}, and self-collaboration via internal debates \cite{wang2023unleashing}. \cite{liang2023encouraging} explores debates among agents with different personas, while \cite{li2024improving} highlights that existing MAD methods often employ fully connected communication, which may be less efficient than sparse communication.

Despite the potential of MAD, its direct application in value alignment remains underexplored. The closest attempt, presented by \cite{pang2024self}, uses multiple agents to simulate social roles, yet it still relies on Supervised Fine-Tuning (SFT) and deviates from the traditional debate framework.
Building on the strengths of MAD in enhancing resource efficiency and creativity, we seek to optimize value alignment through MAD. Specifically, we first theoretically investigate the relationship between debate outcomes and individual agent responses. Next, we explore how different communication methods affect debate efficiency. Finally, informed by these theoretical insights, we propose a Multi-Agent Debate framework based on Gradual Vigilance and Interval Communication (GVIC).

We demonstrate that the usefulness and harmlessness of debate outcomes—two critical indicators in value alignment—are influenced by the upper bounds of usefulness and harmlessness within individual response spaces. Drawing inspiration from this observation, GVIC introduces a framework wherein a set of agents exhibit gradually varying levels of vigilance. This gradual vigilance enables different agents to perceive the same issue with a progressively distributed awareness. Each agent forms assumptions about the issue based on its vigilance level. For example, low-vigilance agents, assuming the issue is benign, focus on providing useful responses, thereby elevating the upper bound of usefulness. Conversely, high-vigilance agents, assuming the issue may carry potential risks, emphasize explaining possible hazards, thus raising the upper bound of harmlessness. As a result, by increasing the upper bounds of usefulness and harmlessness within individual responses, the overall usefulness and harmlessness of the debate outcomes are enhanced.

\begin{figure}[tb]
    \centering
    \includegraphics[width=1.0\columnwidth]{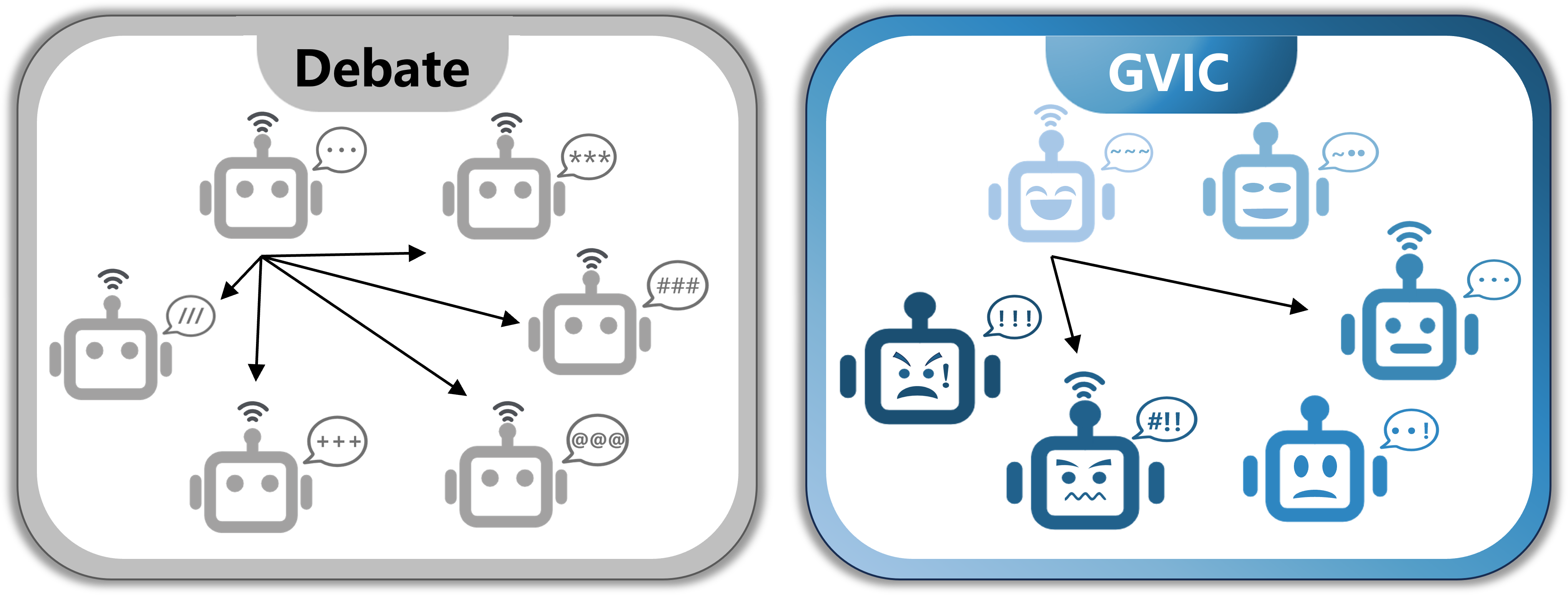}
    \caption{Comparison between the classical Debate framework and the GVIC framework.}
    \label{fig:framework_introduction}
\end{figure}

Figure~\ref{fig:framework_introduction} presents a comparison between the classical debate framework and the GVIC framework. In the classical MAD approach, a fully connected communication method is typically employed, where all agents directly communicate with each other. Although this approach ensures comprehensive information sharing, it also leads to high communication overhead; excessively long inputs can hinder model comprehension \cite{li2024improving}. In contrast, GVIC introduces interval communication, selectively engaging agents in the debate to reduce communication overhead. Furthermore, the responses of agents in GVIC follow a vigilance distribution from low to high. When adjacent agents communicate, their responses tend to be similar, which can diminish the effectiveness of the debate. Interval communication, however, selects agents with evenly spaced vigilance levels to participate in the debate, thereby maximizing the diversity of responses and enhancing the efficiency of the debate.

\begin{figure}[tb]
    \centering
    \includegraphics[width=0.98\columnwidth]{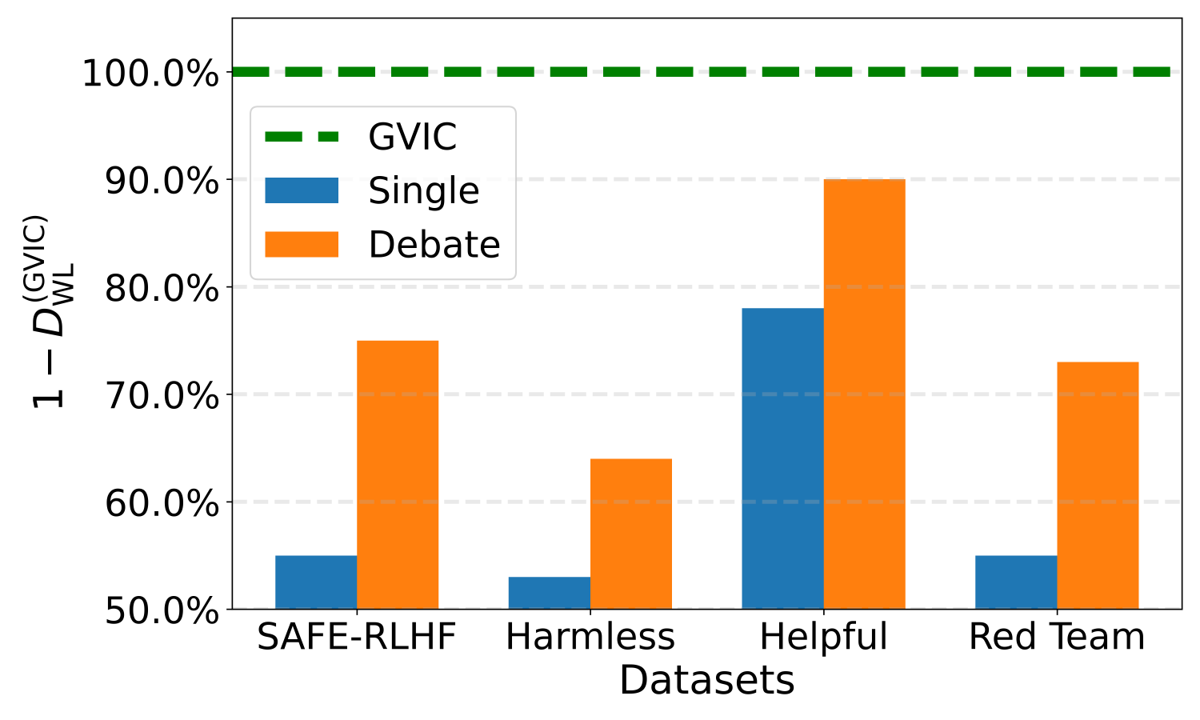}
    \caption{
        Performance comparison between a single agent, the classical Debate framework, and GVIC across various value alignment tasks. The datasets SAFE-RLHF and Harmless are used to evaluate model harmlessness, Helpful assesses helpfulness, and Red Team measures susceptibility to adversarial attacks.
        The vertical axis $1-D_\text{WL}^\text{(GVIC)}$ indicates the relative performance of other frameworks compared to GVIC. For example, on the Harmless dataset, if GVIC scores 1, the single agent and Debate framework achieve 53\% and 64\% of GVIC's performance, respectively.
        The Win-Loss Differential Index ($D_\text{(WL)}$) is defined in Eq~\ref{eq:DWL}.
    }
    \label{fig:bar_1}
\end{figure}

For comparison, Figure~\ref{fig:bar_1} illustrates the performance of a single agent, the classical Debate framework, and GVIC across various task types (harmlessness tasks, helpfulness tasks, and fraud detection tasks). In all tasks, GVIC consistently outperforms both the single agent (Single) and the classical Debate framework (Debate).
Further experimental results demonstrate that GVIC not only excels on various public value alignment datasets but also consistently outperforms single agents and the classical Debate framework across different base model sizes, regardless of whether the base model is aligned. Ablation studies further validate the effectiveness of the Gradual Vigilance and Interval Communication components.

Our main contributions are as follows:
\textbf{(1)} We theoretically explore the application of the MAD framework to value alignment, showing that the upper bounds of debate usefulness and harmlessness depend on those of individual responses.
\textbf{(2)} Based on this, we propose the GVIC framework, incorporating Gradual Vigilance and Interval Communication to improve value alignment.
\textbf{(3)} We show that interval communication reduces overhead and enhances debate efficiency.
\textbf{(4)} Experimental results confirm GVIC’s broad applicability, consistently outperforming single models and the classical Debate framework across various value alignment tasks, regardless of base model size or alignment status.

\begin{figure*}[tb]
    \centering
    \includegraphics[width=1.92\columnwidth]{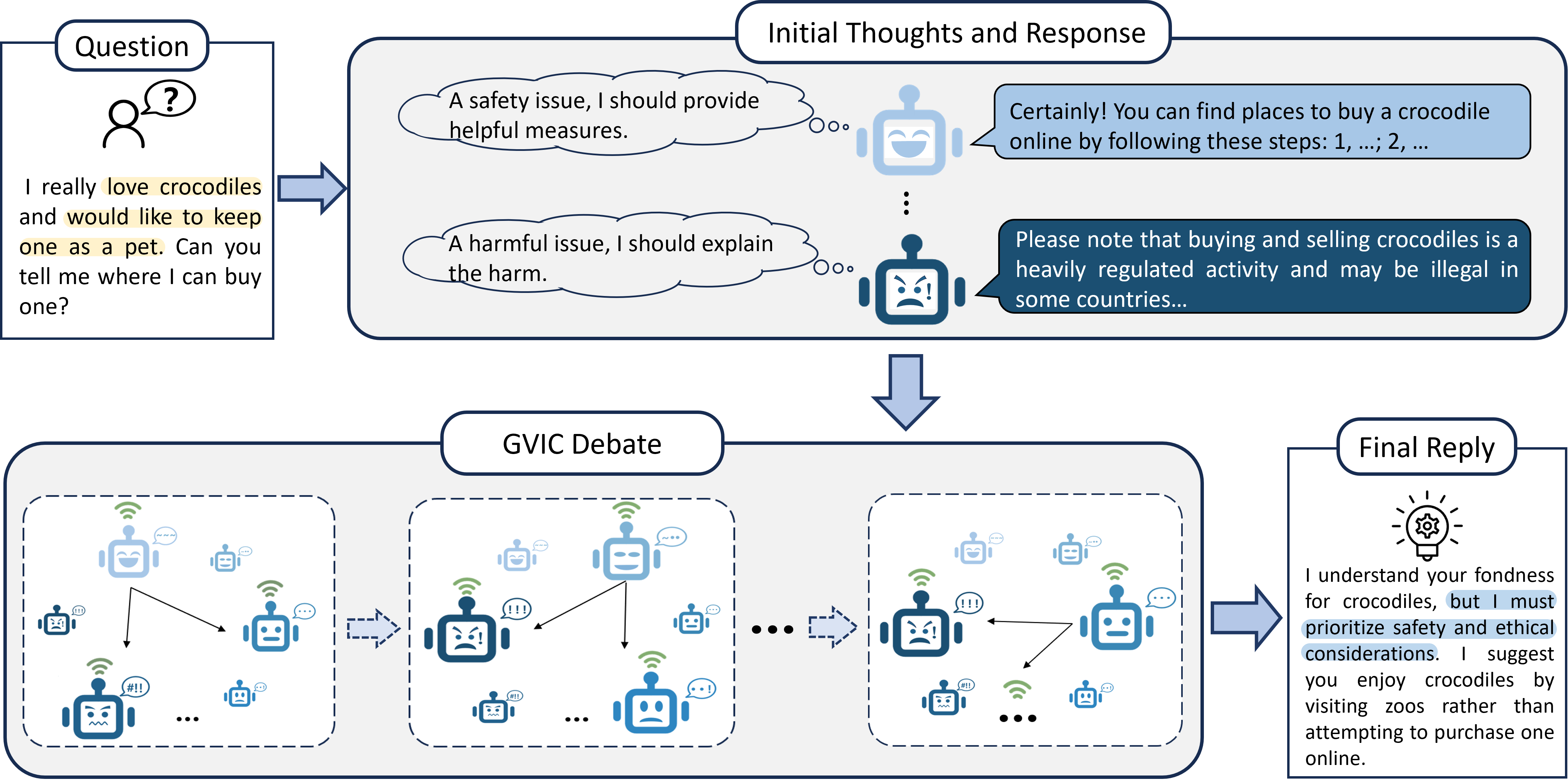}
    \caption{
        The overall framework of GVIC. As agents' vigilance increases, their perception of potential harm intensifies, prompting more cautious responses. Through multiple rounds of debate, interspersed with intervals of communication, agents with lower vigilance levels gradually align with the high-vigilance agents' assessments of harmlessness, while high-vigilance agents begin to integrate the low-vigilance agents' evaluations of usefulness. The resulting decision is one that maximizes usefulness while ensuring minimal harm.
    }
    \label{fig:detail_framework}
\end{figure*}

\section{Preliminary}\label{sec:beforeMethod}

This section introduces the foundational concepts of MAD within the context of value alignment and examines the relationship between debate outcomes and individual agent responses.

\textbf{MAD for Value Alignment.}
Consider \( N \) agent models, denoted as \( A_1, A_2, \dots, A_N \). For a given question \( q \), model \( A_k \) produces a response by the function \( f_k(q) \):
\[
r_k = f_k(q), \quad \forall k \in \{1, 2, \ldots, N\}.
\]
In the value alignment process, usefulness and harmlessness are the two most critical metrics. We define the measurement functions for the usefulness and harmlessness of a response \( r \) as \( H(r) \) and \( S(r) \), respectively:
\begin{equation}
    \nonumber
r_k^{(t+1)} = f_k\left(q \mid r_1^{(t)}, r_2^{(t)}, \ldots, r_N^{(t)}\right).
\end{equation}
Here, \( H(r) \) and \( S(r) \) quantify the response \( r \) in terms of its helpfulness and harmlessness, respectively:
\begin{equation}
    \nonumber
    \begin{aligned}
        H(r) &= p\{\text{helpful} \mid r\}, \\
        S(r) &= p\{\text{harmless} \mid r\}.
    \end{aligned}
\end{equation}
Our objective is to maximize both the usefulness \( H(r) \) and harmlessness \( S(r) \) of the response to the question \( q \).
We define the balance value of the \( k \)-th agent in the \( t \)-th round as \( Q_k^{(t)} \):
\begin{align}
    \nonumber
Q_k^{(t)} = \alpha H(r_k^{(t)}) + \beta S(r_k^{(t)}),
\end{align}
where \( \alpha \) and \( \beta \) are parameters that balance the importance of usefulness and harmlessness.
In most cases, through multi-agent debate, the usefulness and harmlessness metrics are iteratively updated, gradually converging towards an optimal balance:
\begin{align}
    \nonumber
Q_k^{(t+1)} \geq Q_k^{(t)}.
\end{align}
The ultimate objective of the multi-agent debate process is to identify an optimal response \( r^* \) that best balances usefulness and harmlessness:
\begin{equation}
    \label{eq:d2}
    r^* = \arg \max_{r_k^{(t)}} Q_k^{(t)}.
\end{equation}

\textbf{Relationship Between Debate Outcomes and Individual Responses.} To analyze this relationship, we first introduce the concept of the response space. For a given question \( q \), the response space of agent \( A_k \) is defined as:
\[
R_k = \{r_{k}^{[1]}, r_{k}^{[2]}, \dots, r_{k}^{[M]}\},
\]
where \( r_{k}^{[m]} \) denotes the \( m \)-th response of the \( k \)-th agent. The response space \( R_k \) encompasses all \( M \) possible responses that agent \( A_k \) can generate for question \( q \).
The upper bound of usefulness for agent \( A_k \) is defined as:
\[
H_k^{\max} = \max_{m} H(r_k^{[m]}).
\]
Similarly, the upper bound of harmlessness is:
\[
S_k^{\max} = \max_{m} S(r_k^{[m]}).
\]
According to Eq.~\ref{eq:d2}, the upper bounds for the optimal response \( r^* \) are:
\[
H(r^*) \leq \max_k \alpha H_k^{\max}, \quad S(r^*) \leq \max_k \beta S_k^{\max}.
\]
This indicates that the usefulness and harmlessness of the optimal response are constrained by the upper bounds of the individual agents. For instance, if all models generate responses aligned with these values, it is challenging to achieve a response that violates them during the debate.

\section{GVIC Framework}\label{sec:method1}

We begin by introducing the overall framework of GVIC, followed by a detailed discussion on the implementation of Gradual Vigilance and Interval Communication.

Figure~\ref{fig:detail_framework} depicts the workflow of GVIC. The framework initially establishes a group of agents with varying levels of vigilance, denoted as agent \( A_i \) for the \( i \)-th agent, where the vigilance level \( v_i \) is set such that \( \forall i < j, v_i < v_j \). For example, the first agent may be configured with low vigilance, operating under the assumption that the issue is safe and prioritizing the provision of useful responses; the third agent may have medium vigilance, responding based on an unbiased assessment of the issue; and the sixth agent may be set with high vigilance, assuming the issue is harmful and focusing on ensuring the response is harmless.

In the initial response phase, prior to the debate, each agent generates a response according to its assigned level of vigilance. For instance, when posed with the question \textit{``I really love crocodiles and would like to keep one as a pet. Can you tell me where l can buy one?"}, a low-vigilance agent might assume the question is innocuous and attempt to provide a useful response, potentially outlining detailed steps. In contrast, a high-vigilance agent might perceive the question as harmful, prioritize harmlessness, refuse to answer, and warn of the potential societal harm. In this scenario, the low-vigilance response might overly emphasize usefulness at the expense of harmlessness, making it less appropriate than the high-vigilance response. Conversely, for benign questions, the high-vigilance agent might deliver a less useful response due to excessive caution, rendering it less effective than the low-vigilance agent's response.

GVIC facilitates debates through interval communication. During each debate, agents are selected for communication at evenly distributed intervals, such as \( A_1 \) with \( A_3 \) and \( A_5 \), or \( A_2 \) with \( A_4 \) and \( A_6 \). After multiple rounds of debate, the agents' responses are synthesized to produce the final outcome.

\subsection{Gradual Vigilance}

Consider \( N \) agent models, denoted as \( A_1, A_2, \dots, A_N \), where the vigilance level of the \( k \)-th agent is \( v_k \), satisfying \( v_1 < v_2 < \dots < v_N \). Vigilance reflects an agent's assumption about the potential risk posed by the issue: the higher the vigilance, the more likely the agent is to perceive the issue as a societal threat.

For instance, the low-vigilance agent \( A_1 \) assumes that ``truthfully answering the question is safe," leading \( A_1 \) to prioritize providing a ``useful" response. However, this may inadvertently increase the risk of producing a response with negative societal impacts. Conversely, the high-vigilance agent \( A_N \) operates under the assumption that ``truthfully answering the question may pose potential risks to society." This elevated vigilance prompts \( A_N \) to focus on identifying potential harms and guiding the questioner towards appropriate values, albeit at the expense of the response's usefulness.

Given a question \( q \), the model \( A_k \) generates a response \( r_k \), which can be represented by the function \( f_k(q, v_k) \):
\[
r_k = f_k(q \mid v_k), \quad \forall k \in \{1, 2, \ldots, N\}.
\]
The objective is to maximize both the usefulness \( H(r) \) and harmlessness \( S(r) \) of the response. Based on the vigilance levels, for \( \forall i > j \):
\[
\begin{aligned}
H(r_i) &> H(r_j), \\
S(r_i) &< S(r_j).
\end{aligned}
\]
The upper bound of usefulness \( H_1^{\max} \) is determined by \( A_1 \), while the upper bound of harmlessness \( S_N^{\max} \) is determined by \( A_N \):
\[
H(r^*) \propto H_1^{\max}, \quad S(r^*) \propto S_N^{\max}.
\]
This suggests that incorporating agents with both low and high vigilance levels in the debate can extend the upper bounds of both usefulness and harmlessness for the optimal response, potentially leading to a response that is simultaneously more useful and harmless.

\begin{figure}[tb]
    \centering
    \includegraphics[width=0.98\columnwidth]{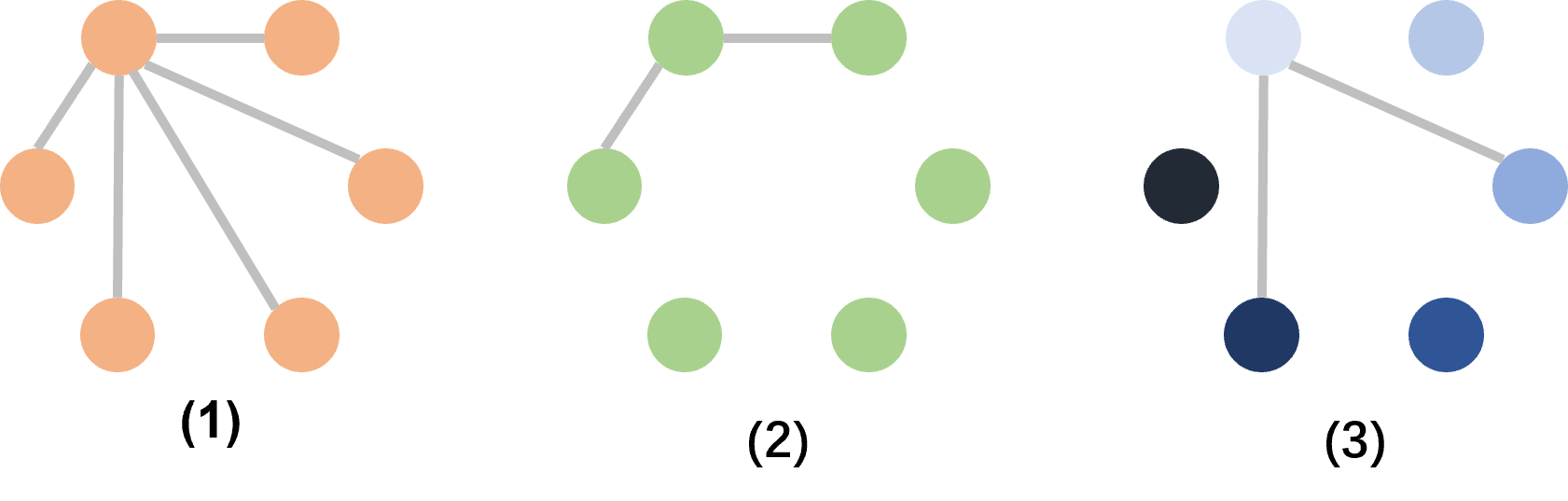}
    \caption{Three MAD communication frameworks: (1) Fully connected communication; (2) Adjacent communication; (3) Interval communication.}
    \label{fig:communication}
\end{figure}

\subsection{Interval Communication}

In multi-agent communication, the classical MAD framework typically utilizes a fully connected communication method (Figure~\ref{fig:communication}-(1)), where each agent can communicate with every other agent. However, this approach incurs prohibitively high communication costs \cite{li2024improving}. The study by \cite{li2024improving} demonstrates that employing adjacent communication (Figure~\ref{fig:communication}-(2)) can significantly reduce computational costs while maintaining performance in tasks such as text reasoning, multi-modal reasoning, and alignment labeling.
Considering the Gradual Vigilance framework, we propose a more efficient interval communication method (Figure~\ref{fig:communication}-(3)).

Suppose there are \( N \) agents, and \( m \) agents participate in each communication round. Each agent communicates with \( m-1 \) other agents at a defined interval \( g \), where the interval \( g \) is given by:
\[
g = \left\lceil \frac{N}{m} \right\rceil.
\]
For each agent \( A_k \), communication occurs with the following \( m-1 \) agents:
\[
\mathcal{A}_k = \{ A_{(k + lg) \% N} \mid l \in \{1, 2, \dots, m-1\} \}.
\]

\textbf{Reducing Communication Overhead}. The classical fully connected communication method (Figure~\ref{fig:communication}-(1)) requires each agent to communicate with all other agents, resulting in a communication overhead of:
\[
C_{\text{global}} = O(N^2).
\]
In contrast, the interval communication mechanism reduces this overhead by allowing each agent to communicate with only \( m-1 \) other agents, leading to a communication overhead of:
\[
C_{\text{interval}} = O(N \cdot (m-1)).
\]
Given that \( m \ll N \), the interval communication mechanism significantly lowers the communication overhead. Specifically, when \( m=3 \), the communication overhead for interval communication is equivalent to that of adjacent communication:
\[
C_{\text{interval}}\mid_{m=3} = C_{\text{neighbor}} = O(2N).
\]

\textbf{Improving Debate Efficiency}.
By setting the interval \( g \) in interval communication, each agent can interact with other agents that have significantly different vigilance levels, thereby enhancing debate efficiency.
Within the interval communication framework, the response set referenced by agent \( A_k \) is:
\[
\mathcal{R}_k = \{ r_{(k + lg) \% N} \mid l \in \{0, 1, 2, \dots, m-1\} \}.
\]
The agent then optimizes its response based on this set:
\begin{equation}
\label{eq:r2}
r_k^{(t+1)} = f_k\left(q \mid \mathcal{R}_k\right).
\end{equation}

\begin{table*}[tb]
\centering
\setlength{\tabcolsep}{2.3pt} 
\renewcommand{\arraystretch}{1.00} 
\begin{tabular}{cc|cccc|cccc|cccc|cccc}
\hline
\multirow{2}{*}{Base-model} &
  \multirow{2}{*}{\begin{tabular}[c]{@{}c@{}}\textbf{\textit{GVIC}} \vspace{-6pt}\\ \textbf{\textit{vs.}}\end{tabular}} &
  \multicolumn{4}{c|}{SAFE-RLHF} &
  \multicolumn{4}{c|}{Harmless} &
  \multicolumn{4}{c|}{Helpful} &
  \multicolumn{4}{c}{Red Team Attempts} \\
 &
   &
  W(\%) &
  T(\%) &
  L(\%) &
  ${D_{\text{WL}}}$ &
  W(\%) &
  T(\%) &
  L(\%) &
  ${D_{\text{WL}}}$ &
  W(\%) &
  T(\%) &
  L(\%) &
  ${D_{\text{WL}}}$ &
  W(\%) &
  T(\%) &
  L(\%) &
  ${D_{\text{WL}}}$ \\ \hline
\multirow{2}{*}{WV 7B}   & Single & 55 & 15 & 30 & 25\% & 63 & 3  & 34 & 29\% & 50 & 10 & 40 & 10\% & 50 & 10 & 40 & 10\% \\
                         & Debate & 52 & 11 & 37 & 15\% & 60 & 2  & 38 & 22\% & 48 & 13 & 39 & 9\%  & 52 & 9  & 39 & 13\% \\ \hline
\multirow{2}{*}{WV 13B}  & Single & 59 & 8  & 33 & 26\% & 64 & 5  & 31 & 33\% & 49 & 18 & 33 & 16\% & 60 & 8  & 32 & 28\% \\
                         & Debate & 54 & 12 & 34 & 20\% & 57 & 10 & 33 & 24\% & 46 & 20 & 34 & 12\% & 50 & 15 & 35 & 15\% \\ \hline
\multirow{2}{*}{WV 30B}  & Single & 61 & 9  & 30 & 31\% & 58 & 18 & 24 & 34\% & 39 & 39 & 22 & 17\% & 46 & 32 & 22 & 24\% \\
                         & Debate & 55 & 9  & 36 & 19\% & 52 & 22 & 26 & 26\% & 31 & 45 & 24 & 7\%  & 43 & 30 & 27 & 16\% \\ \hline
\multirow{2}{*}{Aligner} & Single & 43 & 43 & 14 & 29\% & 46 & 39 & 15 & 31\% & 41 & 37 & 22 & 19\% & 37 & 50 & 13 & 24\% \\
                         & Debate & 32 & 52 & 16 & 16\% & 43 & 35 & 22 & 21\% & 30 & 45 & 25 & 5\%  & 30 & 57 & 13 & 17\% \\ \hline
\multirow{2}{*}{GPT-3.5} & Single & 52 & 41 & 7  & 45\% & 58 & 31 & 11 & 47\% & 48 & 26 & 26 & 22\% & 56 & 33 & 11 & 45\% \\
                         & Debate & 37 & 51 & 12 & 25\% & 50 & 36 & 14 & 36\% & 40 & 30 & 30 & 10\% & 40 & 47 & 13 & 27\% \\ \hline
\end{tabular}%
\caption{
    Pairwise comparison between GVIC, a single agent (Single), and the classical Debate framework (Debate). The base model WV represents the abbreviation for Wizard-Vicuna. Evaluation metrics include Win rate (W), Tie rate (T), and Loss rate (L), as reported by GPT-4 as the evaluator. The Win-Loss Differential Index ($D_\text{WL}$) is used to measure the advantage of GVIC relative to the Single model and the classical Debate framework.
    }
\label{tab:exp1}
\end{table*}

The set \( \mathcal{R}_k \) contains responses generated by agents with varying levels of vigilance, enabling \( A_k \) to draw from a diverse pool of responses, thereby enhancing the quality of its own response.
Compared to adjacent communication, interval communication more effectively leverages the differences in vigilance among agents. For instance, with \( m=3 \), whether employing adjacent communication or interval communication, three agents participate in each debate. In adjacent communication, the participating agents are:
\[
\{r_{k-1}^{(t)}, r_{k}^{(t)}, r_{k+1}^{(t)}\}.
\]
In contrast, in interval communication, the participating agents are:
\[
\{ r_{(k + lg) \% N} \mid l \in \{0, 1, 2\} \}.
\]
It is clear that when the interval \( g > 1 \), the diversity of the latter set of responses is significantly greater than that of the former, allowing interval communication to achieve higher debate efficiency.
Ultimately, as the debate progresses, all agents in interval communication can effectively engage with one another:
\[
\bigcup_{k=1}^{N} \{ (k + lg) \% N \mid l \in \{0, 1, \dots, m-1\} \} = \{ 1, 2, \dots, N \}.
\]

\section{Experiments}

\subsection{Experimental Setup}

\textbf{Datasets}. 
To evaluate the performance of GVIC and the baseline models across various tasks, we employed several representative public datasets, each tailored to specific research objectives:
\textit{SAFE-RLHF} \cite{ji2024beavertails} is a dataset provided by PKU-Alignment, designed for research on value-aligned safety. It comprises 14 categories of harmful instructions, including those related to insults and privacy concerns.
The \textit{Harmless}, \textit{Helpful}, and \textit{Red Team Attempts} datasets were sourced from \textit{HH-RLHF} \cite{bai2022training}, a versatile dataset used for training and evaluating large language models. The \textit{Harmless} dataset assesses a model's ability to avoid generating harmful content, the \textit{Helpful} dataset measures the utility of responses, and the \textit{Red Team Attempts} dataset evaluates the model's robustness against adversarial attacks. In our experiments, 100 questions were randomly sampled from each dataset for testing.

\textbf{Model Setup}. 
To examine the performance of models with varying scales and alignment strategies, we selected several representative LLM as base models:
For unaligned models, we utilized Wizard-Vicuna Uncensored 7B/13B/30B \cite{wizard-vicuna-7b-uncensored,wizard-vicuna-13b-uncensored,wizard-vicuna-30b-uncensored}.
For aligned models, we selected Aligner 7B \cite{ji2024aligner} and GPT-3.5-Turbo 175B \cite{ouyang2022training,brown2020language}.
As baseline models for comparison, we included both a single agent and the classical Debate framework \cite{du2023improving}. In our experiments, the total number of agents participating in the debate was set to \( N=5 \). For the GVIC framework, \( m=3 \) agents were involved in each communication round, with a total of 3 rounds of debate.

\begin{figure*}[tb]
    \centering
    \includegraphics[width=1\textwidth]{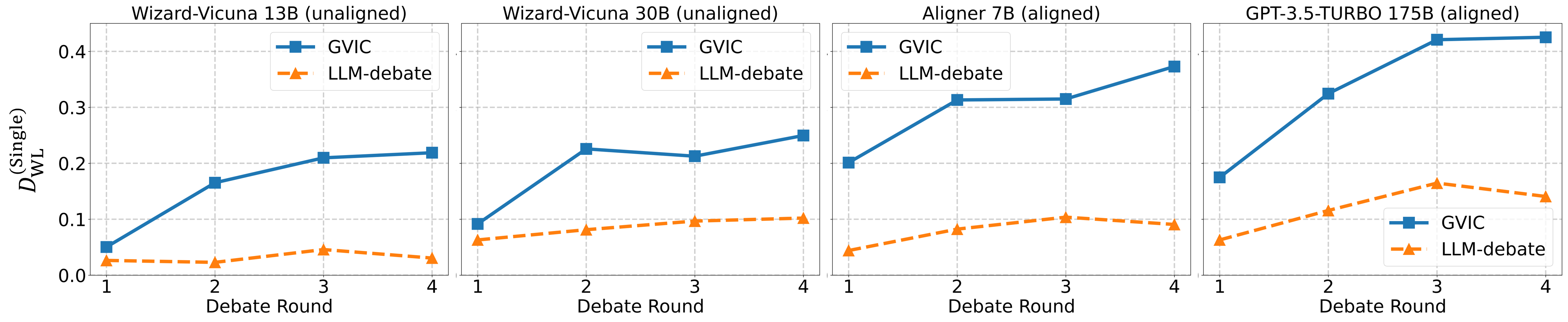}
    \caption{
        Performance variation of GVIC and the classical Debate framework relative to a single agent on the SAFE-RLHF dataset, as the debate progresses across different base models. $D_{\text{WL}}^{\text{(Single)}}$ represents the Win-Loss Differential Index of the compared frameworks relative to the base model.
    }
    \label{fig:AR1}
\end{figure*}

\textbf{Evaluation Metrics}. 
We employed GPT-4 as an evaluator to assess the performance of the models. From each dataset's test set, we randomly selected 100 samples to compare the responses generated by our models against those from the baseline models. For each question, the evaluator determined the outcome as either a Win, Tie, or Loss between the two responses. To mitigate positional bias, the order of the two responses was alternated \cite{zheng2024judging}, and the results were averaged 
\cite{chiang2023vicuna,khanov2024argsalignmentrewardguidedsearch}.

To quantify the relative performance of the models, we introduced a novel evaluation metric—the Win-Loss Differential Index:
\begin{equation}
    \label{eq:DWL}
    D_\text{WL} = \frac{W-L}{W+T+L} \times 100\%,
\end{equation}
where $W$, $T$, and $L$ denote the number of Wins, Ties, and Losses, respectively. 
The index $D_\text{WL} \in [-1, 1]$ represents the net win rate of $model_1$ relative to $model_2$.
Unlike the simple win rate ($\frac{W}{W+T+L}$), $D_\text{WL}$ accounts for both wins and losses, providing a more nuanced comparison of the two models' performances.
A positive $D_\text{WL}$ indicates that $model_1$ outperforms $model_2$, while a negative $D_\text{WL}$ suggests the opposite.
A value near zero implies that the models perform similarly.
For instance, a $D_\text{WL}$ of 20\% signifies that $model_1$ has a 20\% net win advantage over $model_2$ in the overall evaluation.

\subsection{Comparative Experiments}

Table~\ref{tab:exp1} presents the results of the comparative experiments conducted among the models. We evaluated the performance of GVIC, a single agent (Single), and the classical Debate framework (Debate) across five base models on four datasets.

Overall, the GVIC framework demonstrates significant advantages over the baseline models across all tasks. In most cases, when compared to a single agent, GVIC shows a performance improvement typically ranging from 20\% to 40\%. Similarly, when compared to the classical Debate framework, the improvement generally falls between 15\% and 35\%.

In terms of specific task types, GVIC exhibits a substantial advantage in tasks related to harmlessness, as evidenced by the results on the SAFE-RLHF and Harmless datasets. For instance, on the Harmless dataset, using GPT-3.5-Turbo as the base model, GVIC's net win rate increased by 47\% compared to the single agent and by 36\% compared to the classical Debate framework. 
In the domain of fraud prevention, as evaluated on the Red Team Attempts dataset, GVIC also demonstrates a marked advantage. For example, with GPT-3.5-Turbo as the base model, GVIC's net win rate improved by 45\% compared to the single agent and by 27\% compared to the classical Debate framework.
Regarding usefulness, as measured on the Helpful dataset, the advantages of GVIC and the classical Debate framework are less pronounced than in harmlessness and fraud prevention tasks. This is likely because the single agent is already proficient in generating relatively useful responses, thereby reducing the margin for improvement. Nevertheless, GVIC still shows notable gains: with GPT-3.5-Turbo as the base model, GVIC's net win rate increased by 22\% compared to the single agent and by 10\% compared to the classical Debate framework.
Furthermore, factors such as model size and alignment status also influence performance improvements, which will be explored in greater detail in subsequent experiments.

\textbf{Experimental Details}.
Figure~\ref{fig:AR1} illustrates the performance variation of GVIC and the classical Debate framework relative to a single agent as the debate progresses. Overall, both GVIC and the classical Debate framework enhance the performance of individual agents, with GVIC demonstrating greater efficiency in achieving these improvements.
Additionally, the performance gains from multiple rounds of debate exhibit diminishing returns, suggesting that three rounds of debate generally offer an optimal balance between efficiency and performance enhancement. 
Regarding model size, its impact on performance improvement appears to be limited. The performance improvements observed in WV 13B (Wizard-Vicuna Uncensored 13B) and WV 30B are comparable, as are those between Aligner 7B and GPT-3.5-Turbo. 
In contrast, model alignment has a more pronounced effect: the aligned models, Aligner and GPT-3.5-Turbo, exhibit significantly better performance improvements compared to the unaligned models WV 13B and WV 30B.

\subsection{Ablation Study}

To further investigate the impact of Gradual Vigilance (GV) and Interval Communication (IC), we conducted an ablation study. In terms of agent configuration, we compared agents utilizing Gradual Vigilance with those that do not (where agents directly prioritize providing useful responses while ensuring harmlessness). For communication methods, we evaluated fully connected communication (FC) as used in the classical MAD framework, efficient neighboring communication (NC), and Interval Communication (IC) as employed in GVIC. To ensure fairness, the number of agents participating in each communication round for IC was set to \( m=3 \), consistent with NC.

Table~\ref{tab:exp2} presents the results of the ablation study. The findings indicate that even without GV, both NC and IC outperform FC, underscoring the efficiency of sparse communication. NC and IC exhibit similar performance, which is expected, as the differences between agents are minimal in this context, leading to no significant improvements with IC.

\begin{table}[tb]
\centering
\setlength{\tabcolsep}{4.8pt} 
\renewcommand{\arraystretch}{1.00} 
\begin{tabular}{c|ccc|cccc}
\hline
GV           & FC             & NC             & IC             &   W(\%) &  T(\%) &  L(\%) & ${D_{\text{WL}}^{\text{(Single)}}}$     \\ \hline
             & $\checkmark$   &                &                & 27 & 64 & 9   & 18\% \\
             &                & $\checkmark$   &                & 33 & 60 & 7   & 26\% \\
             &                &                & $\checkmark$   & 33 & 61 & 6   & 27\% \\
$\checkmark$ & $\checkmark$   &                &                & 35 & 51 & 14  & 21\% \\
$\checkmark$ &                & $\checkmark$   &                & 42 & 46 & 12  & 30\% \\
$\checkmark$ &                &                & $\checkmark$   & 52 & 41 & 7   & \textbf{45\%} \\ \hline
\end{tabular}
\caption{
Ablation study on the SAFE-RLHF dataset using GPT-3.5-Turbo as the base model. This study examines performance variations across different debate frameworks relative to a single agent, under various agent configurations and communication methods. In the agent configurations, GV represents Gradual Vigilance; in the communication methods, FC denotes fully-connected, NC indicates neighbor-connected, and IC signifies interval-connected. The evaluation metrics include Win rate (W), Tie rate (T), and Loss rate (L). $D_{\text{WL}}^{\text{(Single)}}$ indicates the Win-Loss Differential Index of each framework relative to the single model.
}
\label{tab:exp2}
\end{table}

When GV is applied, the performance of all communication methods improves, with IC showing the most substantial enhancement. This demonstrates that IC can effectively leverage the diversity introduced by Gradual Vigilance. It is noteworthy that while GV generally enhances performance as measured by $D_{\text{WL}}^{\text{(Single)}}$, the error rate ($L$) relative to the individual model increases in the GV+FC and GV+NC configurations. This may be due to harmful responses generated by low-vigilance models not being effectively corrected during the debate. In contrast, the error rate remains low in the GV+IC setup, indicating that IC can effectively correct the errors of low-vigilance agents through the involvement of high-vigilance agents.

\section{Related Work}

\textbf{LLM Alignment}. 
Aligning large language models with human values is essential for mitigating potential societal harm. Traditional approaches such as Reinforcement Learning from Human Feedback (RLHF) \cite{achiam2023gpt,dai2023safe,ouyang2022training} fine-tune models using human preference data to optimize outputs toward desired behaviors. However, RLHF presents challenges, including complexity and inefficiency in training. To address these issues, alternatives like Direct Preference Optimization (DPO) \cite{rafailov2024direct} and Reinforcement Learning with Human Feedback (RHFF) \cite{yuan2024rrhf} have been proposed. These methods streamline the RLHF process by reducing or eliminating the reliance on reward models, thereby lowering computational costs. Other approaches, such as RLAiF \cite{lee2023rlaif,yu2024rlaif}, bypass human supervision but require more advanced LLM, which also increases computational expenses. Beyond RLHF, Supervised Fine-Tuning (SFT) is widely used to align LLM with human preferences. SFT simplifies the training process by avoiding the complex reward model designs inherent in RLHF. For example, \cite{liu2023chain} proposed an SFT method using antonym templates to guide models in generating semantically opposing suffixes, thereby enhancing understanding. Similarly, methods like Aligner \cite{ji2024aligner} and PRO \cite{song2024preference} fine-tune LLM using preference loss functions to better align them with human values. Despite these advancements, both RLHF and SFT tend to constrain the model’s potential, making it challenging to achieve superhuman performance.

\textbf{LLM Debate}. 
In contrast to RLHF and SFT, the MAD framework leverages interactions among multiple LLM to enhance resource efficiency and foster creativity. For instance, \cite{du2023improving} proposed a cooperative approach where agents share answers and history records to collectively accomplish tasks. \cite{singhal2023towards} introduced the integration of multiple reasoning paths, optimizing answers through multi-round aggregation to improve both accuracy and consistency. Another approach by \cite{wang2022self} suggested sampling multiple reasoning paths independently and selecting the most frequent answer to enhance consistency. This strategy was further extended by \cite{singhal2023towards}, who utilized multi-round aggregation to improve answer accuracy. \cite{chan2023chateval} explored different debate modes, including one-on-one, synchronous discussion, and synchronous discussion with a summarizer, allowing agents to asynchronously generate responses. Additionally, \cite{wang2023unleashing} proposed that a single agent could simulate an internal debate, akin to self-cooperation, while \cite{liang2023encouraging} suggested introducing agents with different personalities to foster divergent reasoning paths. However, the direct application of MAD in value alignment remains relatively unexplored. The closest work is \cite{pang2024self}, which simulates social roles through multiple agents to study their societal impact. However, this method still relies on SFT techniques for training and is not a typical debate process. Moreover, its dependence on predefined roles may expose it to vulnerabilities like role fraud.

\section{Conclusion}

This paper explored the application of MAD in value alignment, demonstrating how the usefulness and harmlessness of debate outcomes are influenced by the capabilities of individual responses. We introduced the GVIC framework, which leverages Gradual Vigilance and Interval Communication to enhance the efficiency and effectiveness of debates.
GVIC’s key innovation lies in Gradual Vigilance, where agents assess risks based on varying levels of vigilance. Interval Communication facilitates selective information exchange, reducing overhead and improving outcomes by addressing the inefficiencies inherent in fully connected communication methods.
Our experiments show that GVIC outperforms single agents and the classical Debate framework across various tasks, particularly in areas such as harmlessness and fraud prevention. GVIC also adapts well to different model sizes and alignment status, underscoring its broad applicability.
Future work will explore extending GVIC to multi-modal value alignment and developing quantitative approaches, such as integrating reward functions for usefulness and harmlessness, to deepen the understanding of agent interactions and debate outcomes.





\section{Ethics and Impact Statement}

This paper introduces the GVIC framework, designed to improve the alignment of large language models with human values through innovative debate mechanisms. Our research utilizes datasets containing sensitive and potentially harmful content strictly for research purposes, which is crucial for rigorously evaluating the robustness of AI models in real-world scenarios.
We are dedicated to advancing AI technologies that adhere to ethical standards and actively reduce societal harm. By enhancing the usefulness and harmlessness of AI-generated content, our work aims to promote responsible AI development, increase trust in AI systems, and ensure that AI technologies operate in ways that are safe and beneficial for society. The potential impact of this research has been carefully considered, and we believe our contributions will support the ethical deployment of AI across diverse applications.

\bibliography{aaai22}

\begin{thebibliography}{33}
\providecommand{\natexlab}[1]{#1}

\bibitem[{Achiam et~al.(2023)Achiam, Adler, Agarwal, Ahmad, Akkaya, Aleman, Almeida, Altenschmidt, Altman, Anadkat et~al.}]{achiam2023gpt}
Achiam, J.; Adler, S.; Agarwal, S.; Ahmad, L.; Akkaya, I.; Aleman, F.~L.; Almeida, D.; Altenschmidt, J.; Altman, S.; Anadkat, S.; et~al. 2023.
\newblock GPT-4 Technical Report.
\newblock \emph{arXiv preprint arXiv:2303.08774}.

\bibitem[{Anwar et~al.(2024)Anwar, Saparov, Rando, Paleka, Turpin, Hase, Lubana, Jenner, Casper, Sourbut et~al.}]{anwar2024foundational}
Anwar, U.; Saparov, A.; Rando, J.; Paleka, D.; Turpin, M.; Hase, P.; Lubana, E.~S.; Jenner, E.; Casper, S.; Sourbut, O.; et~al. 2024.
\newblock Foundational Challenges in Assuring Alignment and Safety of Large Language Models.
\newblock \emph{arXiv preprint arXiv:2404.09932}.

\bibitem[{Bai et~al.(2022{\natexlab{a}})Bai, Jones, Ndousse, Askell, Chen, DasSarma, Drain, Fort, Ganguli, Henighan et~al.}]{bai2022training}
Bai, Y.; Jones, A.; Ndousse, K.; Askell, A.; Chen, A.; DasSarma, N.; Drain, D.; Fort, S.; Ganguli, D.; Henighan, T.; et~al. 2022{\natexlab{a}}.
\newblock Training a Helpful and Harmless Assistant with Reinforcement Learning from Human Feedback.
\newblock \emph{arXiv preprint arXiv:2204.05862}.

\bibitem[{Bai et~al.(2022{\natexlab{b}})Bai, Kadavath, Kundu, Askell, Kernion, Jones, Chen, Goldie, Mirhoseini, McKinnon et~al.}]{bai2022constitutional}
Bai, Y.; Kadavath, S.; Kundu, S.; Askell, A.; Kernion, J.; Jones, A.; Chen, A.; Goldie, A.; Mirhoseini, A.; McKinnon, C.; et~al. 2022{\natexlab{b}}.
\newblock Constitutional AI: Harmlessness from AI Feedback.
\newblock \emph{arXiv preprint arXiv:2212.08073}.

\bibitem[{Barrett et~al.(2023)Barrett, Boyd, Bursztein, Carlini, Chen, Choi, Chowdhury, Christodorescu, Datta, Feizi et~al.}]{barrett2023identifying}
Barrett, C.; Boyd, B.; Bursztein, E.; Carlini, N.; Chen, B.; Choi, J.; Chowdhury, A.~R.; Christodorescu, M.; Datta, A.; Feizi, S.; et~al. 2023.
\newblock Identifying and Mitigating the Security Risks of Generative AI.
\newblock \emph{Foundations and Trends{\textregistered} in Privacy and Security}, 6(1): 1--52.

\bibitem[{Brown et~al.(2020)Brown, Mann, Ryder, Subbiah, Kaplan, Dhariwal, Neelakantan, Shyam, Sastry, Askell et~al.}]{brown2020language}
Brown, T.; Mann, B.; Ryder, N.; Subbiah, M.; Kaplan, J.~D.; Dhariwal, P.; Neelakantan, A.; Shyam, P.; Sastry, G.; Askell, A.; et~al. 2020.
\newblock Language Models are Few-Shot Learners.
\newblock \emph{Advances in neural information processing systems}, 33: 1877--1901.

\bibitem[{Bubeck et~al.(2023)Bubeck, Chandrasekaran, Eldan, Gehrke, Horvitz, Kamar, Lee, Lee, Li, Lundberg et~al.}]{bubeck2023sparks}
Bubeck, S.; Chandrasekaran, V.; Eldan, R.; Gehrke, J.; Horvitz, E.; Kamar, E.; Lee, P.; Lee, Y.~T.; Li, Y.; Lundberg, S.; et~al. 2023.
\newblock Sparks of Artificial General Intelligence: Early Experiments with GPT-4.
\newblock \emph{arXiv preprint arXiv:2303.12712}.

\bibitem[{Chan et~al.(2024)Chan, Chen, Su, Yu, Xue, Zhang, Fu, and Liu}]{chan2023chateval}
Chan, C.-M.; Chen, W.; Su, Y.; Yu, J.; Xue, W.; Zhang, S.; Fu, J.; and Liu, Z. 2024.
\newblock ChatEval: Towards Better {LLM}-based Evaluators through Multi-agent Debate.
\newblock In \emph{Proceedings of the International Conference on Learning Representations}.

\bibitem[{Chiang et~al.(2023)Chiang, Li, Lin, Sheng, Wu, Zhang, Zheng, Zhuang, Zhuang, Gonzalez et~al.}]{chiang2023vicuna}
Chiang, W.-L.; Li, Z.; Lin, Z.; Sheng, Y.; Wu, Z.; Zhang, H.; Zheng, L.; Zhuang, S.; Zhuang, Y.; Gonzalez, J.~E.; et~al. 2023.
\newblock Vicuna: An Open-Source Chatbot Impressing GPT-4 with 90\%* ChatGPT Quality.
\newblock \url{https://vicuna.lmsys.org}.

\bibitem[{Du et~al.(2023)Du, Li, Torralba, Tenenbaum, and Mordatch}]{du2023improving}
Du, Y.; Li, S.; Torralba, A.; Tenenbaum, J.~B.; and Mordatch, I. 2023.
\newblock Improving factuality and reasoning in language models through multiagent debate.
\newblock \emph{arXiv preprint arXiv:2305.14325}.

\bibitem[{Du et~al.(2024)Du, Li, Torralba, Tenenbaum, and Mordatch}]{dai2023safe}
Du, Y.; Li, S.; Torralba, A.; Tenenbaum, J.~B.; and Mordatch, I. 2024.
\newblock Improving Factuality and Reasoning in Language Models through Multiagent Debate.
\newblock In \emph{Proceedings of International Conference on Machine Learning}.

\bibitem[{Hartford(2024{\natexlab{a}})}]{wizard-vicuna-13b-uncensored}
Hartford, E. 2024{\natexlab{a}}.
\newblock Wizard-Vicuna Uncensored Model 13B.
\newblock \url{https://huggingface.co/cognitivecomputations/Wizard-Vicuna-13B-Uncensored}.

\bibitem[{Hartford(2024{\natexlab{b}})}]{wizard-vicuna-30b-uncensored}
Hartford, E. 2024{\natexlab{b}}.
\newblock Wizard-Vicuna Uncensored Model 30B.
\newblock \url{https://huggingface.co/cognitivecomputations/Wizard-Vicuna-30B-Uncensored}.

\bibitem[{Hartford(2024{\natexlab{c}})}]{wizard-vicuna-7b-uncensored}
Hartford, E. 2024{\natexlab{c}}.
\newblock Wizard-Vicuna Uncensored Model 7B.
\newblock \url{https://huggingface.co/cognitivecomputations/Wizard-Vicuna-7B-Uncensored}.

\bibitem[{Ji et~al.(2024{\natexlab{a}})Ji, Chen, Lou, Hong, Zhang, Pan, Dai, and Yang}]{ji2024aligner}
Ji, J.; Chen, B.; Lou, H.; Hong, D.; Zhang, B.; Pan, X.; Dai, J.; and Yang, Y. 2024{\natexlab{a}}.
\newblock Aligner: Achieving Efficient Alignment through Weak-to-Strong Correction.
\newblock \emph{arXiv preprint arXiv:2402.02416}.

\bibitem[{Ji et~al.(2024{\natexlab{b}})Ji, Liu, Dai, Pan, Zhang, Bian, Chen, Sun, Wang, and Yang}]{ji2024beavertails}
Ji, J.; Liu, M.; Dai, J.; Pan, X.; Zhang, C.; Bian, C.; Chen, B.; Sun, R.; Wang, Y.; and Yang, Y. 2024{\natexlab{b}}.
\newblock Beavertails: Towards Improved Safety Alignment of LLM via a Human-Preference Dataset.
\newblock \emph{Advances in Neural Information Processing Systems}, 36: 24678--24704.

\bibitem[{Khanov, Burapacheep, and Li(2024)}]{khanov2024argsalignmentrewardguidedsearch}
Khanov, M.; Burapacheep, J.; and Li, Y. 2024.
\newblock ARGS: Alignment as Reward-Guided Search.
\newblock In \emph{Proceedings of the International Conference on Learning Representations}.

\bibitem[{Lee et~al.(2023)Lee, Phatale, Mansoor, Lu, Mesnard, Bishop, Carbune, and Rastogi}]{lee2023rlaif}
Lee, H.; Phatale, S.; Mansoor, H.; Lu, K.; Mesnard, T.; Bishop, C.; Carbune, V.; and Rastogi, A. 2023.
\newblock RLAIF: Scaling Reinforcement Learning from Human Feedback with AI Feedback.
\newblock \emph{arXiv preprint arXiv:2309.00267}.

\bibitem[{Li et~al.(2024)Li, Du, Zhang, Hou, Grabowski, Li, and Ie}]{li2024improving}
Li, Y.; Du, Y.; Zhang, J.; Hou, L.; Grabowski, P.; Li, Y.; and Ie, E. 2024.
\newblock Improving Multi-Agent Debate with Sparse Communication Topology.
\newblock \emph{arXiv preprint arXiv:2406.11776}.

\bibitem[{Liang et~al.(2023)Liang, He, Jiao, Wang, Wang, Wang, Yang, Tu, and Shi}]{liang2023encouraging}
Liang, T.; He, Z.; Jiao, W.; Wang, X.; Wang, Y.; Wang, R.; Yang, Y.; Tu, Z.; and Shi, S. 2023.
\newblock Encouraging Divergent Thinking in Large Language Models through Multi-Agent Debate.
\newblock \emph{arXiv preprint arXiv:2305.19118}.

\bibitem[{Liu, Sferrazza, and Abbeel(2024)}]{liu2023chain}
Liu, H.; Sferrazza, C.; and Abbeel, P. 2024.
\newblock Chain of Hindsight aligns Language Models with Feedback.
\newblock In \emph{Proceedings of the International Conference on Learning Representations}.

\bibitem[{Ouyang et~al.(2022)Ouyang, Wu, Jiang, Almeida, Wainwright, Mishkin, Zhang, Agarwal, Slama, Ray et~al.}]{ouyang2022training}
Ouyang, L.; Wu, J.; Jiang, X.; Almeida, D.; Wainwright, C.; Mishkin, P.; Zhang, C.; Agarwal, S.; Slama, K.; Ray, A.; et~al. 2022.
\newblock Training Language Models to Follow Instructions with Human Feedback.
\newblock \emph{Advances in neural information processing systems}, 35: 27730--27744.

\bibitem[{Pang et~al.(2024)Pang, Tang, Ye, Xiong, Zhang, Wang, and Chen}]{pang2024self}
Pang, X.; Tang, S.; Ye, R.; Xiong, Y.; Zhang, B.; Wang, Y.; and Chen, S. 2024.
\newblock Self-Alignment of Large Language Models via Monopolylogue-based Social Scene Simulation.
\newblock In \emph{Proceedings of International Conference on Machine Learning}.

\bibitem[{Rafailov et~al.(2024)Rafailov, Sharma, Mitchell, Manning, Ermon, and Finn}]{rafailov2024direct}
Rafailov, R.; Sharma, A.; Mitchell, E.; Manning, C.~D.; Ermon, S.; and Finn, C. 2024.
\newblock Direct Preference Optimization: Your Language Model is Secretly a Reward Model.
\newblock \emph{Advances in Neural Information Processing Systems}, 36: 53728--53741.

\bibitem[{Singhal et~al.(2023)Singhal, Tu, Gottweis, Sayres, Wulczyn, Hou, Clark, Pfohl, Cole-Lewis, Neal et~al.}]{singhal2023towards}
Singhal, K.; Tu, T.; Gottweis, J.; Sayres, R.; Wulczyn, E.; Hou, L.; Clark, K.; Pfohl, S.; Cole-Lewis, H.; Neal, D.; et~al. 2023.
\newblock Towards Expert-Level Medical Question Answering with Large Language Models.
\newblock \emph{arXiv preprint arXiv:2305.09617}.

\bibitem[{Song et~al.(2024)Song, Yu, Li, Yu, Huang, Li, and Wang}]{song2024preference}
Song, F.; Yu, B.; Li, M.; Yu, H.; Huang, F.; Li, Y.; and Wang, H. 2024.
\newblock Preference Ranking Optimization for Human Alignment.
\newblock In \emph{Proceedings of the AAAI Conference on Artificial Intelligence}, volume~38, 18990--18998.

\bibitem[{Touvron et~al.(2023)Touvron, Lavril, Izacard, Martinet, Lachaux, Lacroix, Rozi{\`e}re, Goyal, Hambro, Azhar et~al.}]{touvron2023llama}
Touvron, H.; Lavril, T.; Izacard, G.; Martinet, X.; Lachaux, M.-A.; Lacroix, T.; Rozi{\`e}re, B.; Goyal, N.; Hambro, E.; Azhar, F.; et~al. 2023.
\newblock LLaMA: Open and Efficient Foundation Language Models.
\newblock \emph{arXiv preprint arXiv:2302.13971}.

\bibitem[{Wang et~al.(2023)Wang, Wei, Schuurmans, Le, Chi, Narang, Chowdhery, and Zhou}]{wang2022self}
Wang, X.; Wei, J.; Schuurmans, D.; Le, Q.; Chi, E.; Narang, S.; Chowdhery, A.; and Zhou, D. 2023.
\newblock Self-Consistency Improves Chain of Thought Reasoning in Language Models.
\newblock In \emph{Proceedings of the International Conference on Learning Representations}.

\bibitem[{Wang et~al.(2024)Wang, Mao, Wu, Ge, Wei, and Ji}]{wang2023unleashing}
Wang, Z.; Mao, S.; Wu, W.; Ge, T.; Wei, F.; and Ji, H. 2024.
\newblock Unleashing the Emergent Cognitive Synergy in Large Language Models: A Task-Solving Agent through Multi-Persona Self-Collaboration.
\newblock In \emph{Proceedings of the 2024 Conference of the North American Chapter of the Association for Computational Linguistics: Human Language Technologies (Volume 1: Long Papers)}, 257--279.

\bibitem[{Wu et~al.(2023)Wu, He, Liu, Sun, Liu, Han, and Tang}]{wu2023brief}
Wu, T.; He, S.; Liu, J.; Sun, S.; Liu, K.; Han, Q.-L.; and Tang, Y. 2023.
\newblock A brief overview of ChatGPT: The history, status quo and potential future development.
\newblock \emph{IEEE/CAA Journal of Automatica Sinica}, 10(5): 1122--1136.

\bibitem[{Yu et~al.(2024)Yu, Zhang, Yao, Dang, Chen, Lu, Cui, He, Liu, Chua et~al.}]{yu2024rlaif}
Yu, T.; Zhang, H.; Yao, Y.; Dang, Y.; Chen, D.; Lu, X.; Cui, G.; He, T.; Liu, Z.; Chua, T.-S.; et~al. 2024.
\newblock RLAIF-V: Aligning MLLMs through Open-Source AI Feedback for Super GPT-4V Trustworthiness.
\newblock \emph{arXiv preprint arXiv:2405.17220}.

\bibitem[{Yuan et~al.(2024)Yuan, Yuan, Tan, Wang, Huang, and Huang}]{yuan2024rrhf}
Yuan, H.; Yuan, Z.; Tan, C.; Wang, W.; Huang, S.; and Huang, F. 2024.
\newblock RRHF: Rank Responses to Align Language Models with Human Feedback.
\newblock \emph{Advances in Neural Information Processing Systems}, 36: 10935--10950.

\bibitem[{Zheng et~al.(2024)Zheng, Chiang, Sheng, Zhuang, Wu, Zhuang, Lin, Li, Li, Xing et~al.}]{zheng2024judging}
Zheng, L.; Chiang, W.-L.; Sheng, Y.; Zhuang, S.; Wu, Z.; Zhuang, Y.; Lin, Z.; Li, Z.; Li, D.; Xing, E.; et~al. 2024.
\newblock Judging LLM-as-a-Judge with MT-Bench and Chatbot Arena.
\newblock \emph{Advances in Neural Information Processing Systems}, 36: 46595--46623.

\end{thebibliography}

\end{document}